\documentclass[10pt,english]{article}
\usepackage{subfigure}
\usepackage{graphicx}
\usepackage{float}
\usepackage[T1]{fontenc}
\usepackage[latin9]{inputenc}
\usepackage[margin=1.25in]{geometry}
\usepackage{float}
\usepackage{amsmath}
\usepackage{amsbsy}
\usepackage{setspace}
\usepackage{amssymb}
\usepackage{esint}
\usepackage{cite}
\newfloat{algorithm}{tbp}{loa}
\floatname{algorithm}{Algorithm}
\usepackage{algorithmic}
\usepackage{hyperref}

\newlength\myindent
\makeatletter

\floatstyle{ruled}
\newfloat{algorithm}{tbp}{loa}
\floatname{algorithm}{Algorithm}

\usepackage{babel}

\begin{document}

\title{K-Means Kernel Classifier}

\author{M. Andrecut}


\maketitle
{

\centering Calgary, Alberta, T3G 5Y8, Canada

\centering mircea.andrecut@gmail.com

} 
\bigskip 
\begin{abstract}
We combine K-means clustering with the least-squares kernel classification method. K-means clustering is used to extract a set 
of representative vectors for each class. The least-squares kernel method uses these representative vectors as a training set for the classification task. 
We show that this combination of unsupervised and supervised learning algorithms performs very well, and we illustrate this approach using the MNIST dataset. 

\smallskip 


\end{abstract}

\section{Introduction}

Kernel classifiers are some of the most important supervised machine learning tools \cite{key-1}. 
The kernel methods transform a given non-linear problem into a linear one by using a 
similarity kernel function $\Omega(x,x')$ defined over pairs of input data points $(x,x')$. 
This way, the input data $x$ is mapped into a feature space $\phi(x)$, 
where the inner product $\langle\cdot,\cdot\rangle$ can be calculated with a 
positive definite kernel function satisfying Mercer's condition \cite{key-2}, such that 
the mapping is done implicitly, without the need to explicitly map the data points $\phi(x)$:
\begin{equation}
\Omega(x,x') = \langle \phi(x),\phi(x')\rangle.
\end{equation}
Another important result is the Representer theorem which shows that any non-linear function 
$f$ can be expressed as a linear combination of kernel products evaluated on the 
training data points $\chi = \{x_n|n=1,\dots,N \}$ \cite{key-1}:
\begin{equation}
f(x) = \sum_{n=1}^{N} a_n \Omega(x,x_n).
\end{equation}

Several kernel classification methods exist in the literature, here we consider the least-squares 
support vector machines approach (LS-SVM)\cite{key-3}. Because of its high  
complexity, the LS-SVM is not a suitable candidate for applications with large data sets. 
In a previous work we have discussed several approximation methods using randomized block kernel matrices, 
that significantly reduce the complexity of the problem \cite{key-4}. Here, we extend the previous work with a 
different approach based on the K-means clustering, a popular unsupervised machine leaning method. 
More exactly, we use the K-means algorithm \cite{key-5} to extract a set of representative vectors for each class. 
These representative vectors are then used by the LS-SVM kernel method as the training set 
for the classification task. 

The described K-means LS-SVM approach has a couple of significant advantages over the previously described randomization 
methods: (1) it is extremely robust, since it has a single tuning parameter (the number of support vectors per class), 
and therefore data overfitting is easily avoided; (2) it is very simple to implement comparing to the previously proposed randomization methods. 
We illustrate this approach using the MNIST data set, which is a 
well known benchmark frequently used in machine learning for image classification \cite{key-6}.

\section{Kernel LS-SVM classifier}

Assume that $K$ classes are encoded using the standard basis in the $\mathbb{R}^K$ space. 
Therefore, if $x_i \in  \mathbb{R}^M$ is a sample from the class $C_k$, 
then the corresponding label $y_i\in \mathbb{R}^K$ is encoded by a binary row vector with 1 in the $k$-th position and 0 
in all other positions:
\begin{equation}
x_i \in C_k \; \Leftrightarrow \; y_{ij} = 
\begin{cases}
		1 & \text{if }\; j=k \\
		0 & \text{otherwise}
\end{cases}.	
\end{equation}

Using the Lagrange multipliers method one can show that for the training data 
$\{(x_i,y_i)\vert x_i \in \mathbb{R}^M, y_i \in \mathbb{R}^K, i=1,\dots,N\}$  
and the feature mapping function $\phi(x)$, the LS-SVM is equivalent to solving the 
following optimization problem \cite{key-3,key-4}:
\begin{equation}
W = \text{arg} \min_{W} \Vert \Phi W - Y\Vert^2 + \varepsilon \Vert W \Vert^2, 
\end{equation}
and the corresponding linear system of equations: 
\begin{equation}
\Phi W = Y,
\end{equation}
where $W$ and $Y$ are $(N+1) \times K$ matrices with the columns:
\begin{equation}
w^{(j)} = 
\begin{bmatrix}
    b_j \\
    a_{1j}\\
    \vdots\\
    a_{Nj}
\end{bmatrix}, \;
y^{(j)} \leftarrow 
\begin{bmatrix}
    0 \\
    y_{1j}\\
    \vdots\\
    y_{Nj} 
\end{bmatrix}, \quad j=1,\dots,K,
\end{equation}
and $\Phi$ is the extended kernel matrix:
\begin{equation}
\Phi = 
\begin{bmatrix}
    0 & \; & u^T  \\
    u & \; &\Omega + \varepsilon I 
\end{bmatrix}.
\end{equation}
Here, $I$ is the $N\times N$ identity matrix, $u = [1,\dots,1]^T$ is an $N$ 
dimensional vector with all the components equal to 1, 
$b_j$ is the bias, $a_{nj}$ are the unknown Lagrange multipliers, $y_{nj}$ are the binary classification values, 
$\varepsilon>0$ is the regularization parameter, and $\Omega = [\Omega_{ij}]_{N\times N}$ is the kernel matrix with $\Omega_{ij}=\langle \phi(x_i),\phi(x_j) \rangle$, $i,j=1,...,N$.
Thus, one can see that the complexity of the problem resides in solving a large linear system, if the training data is large. Hence the need for efficient approximation methods. 

Once the system of equations is solved, the classification of any new sample $x$ is easily performed as following:
\begin{equation}
x \in C_k \; \Leftrightarrow \; k = \text{arg} \max_{j=1,\dots,K} g_j(x),
\end{equation}
where $g_j$ are the softmax functions:
\begin{equation}
g_j(x) = \frac{\sum_{n=1}^N \exp \left( \Omega(x, x_n) a_{nj} + b_j \right)}
{\sum_{i=1}^K \sum_{n=1}^N \exp \left(\Omega(x, x_n) a_{ni} + b_i \right)}.
\end{equation}

\section{K-means clustering}

As mentioned in the introduction, we use the K-means algorithm to extract a set of representative vectors for each class, 
which then will be used as the new training for the LS-SVM kernel classification method. 

Let us consider all the training data corresponding to the class $C_k$, which obviously is a sub-set of the whole 
training data set:
\begin{equation}
\{(x^{k}_i,y^{k}_i)\vert x^{k}_i \in \mathbb{R}^M, y^{k}_i \in \mathbb{R}^K, i=1,\dots,N_k \}, 
\end{equation}
where $N_k$ is the number of training samples from the class $C_k$. 
We use K-means to extract $Q < N_k$ representative vectors (cluster centroids) for the class $C_k$, and let use denote this set by:
\begin{equation}
\{(\xi^{k}_q,y^{k}_q)\vert \xi^{k}_q \in \mathbb{R}^M, y^{k}_q \in \mathbb{R}^K, q=1,\dots,Q\}.
\end{equation}
Obviously, we also consider that $y^k_{qj}=1$ for $j=k$, and $y^k_{qj}=0$ otherwise, since these are the representative vectors for the class $C_k$. 

After extracting the representative vectors for each class $C_k, k=1,...,K$, we can use the $KQ$ representative vectors to replace the 
training samples in the LS-SVM classifier. That is, we train the LS-SVM classifier on the representative vectors set:
\begin{equation}
\{(\xi_q,y_q)\vert \xi_q \in \mathbb{R}^M, y_q \in \mathbb{R}^K, q=1,\dots,KQ \},
\end{equation}
where for notation simplicity we dropped the class index $k$. 
Thus, in the LS-SVM system of equations we simply consider $\Omega_{ij}=\langle \phi(\xi_i),\phi(\xi_j) \rangle$, $i,j=1,...,KQ$, 
and the system size becomes $(KQ+1)^2 < (N+1)^2$. Accordingly, the unknown vector $w^{(j)}$ becomes $KQ+1$ dimensional.

Once the new system of equations is solved, the classification of any new sample $x$ is performed as following:
\begin{equation}
x \in C_k \; \Leftrightarrow \; k = \text{arg} \max_{j=1,\dots,K} g_j(x),
\end{equation}
where $g$ is the softmax function.
\begin{equation}
g_j(x) = \frac{\sum_{q=1}^{KQ} \exp \left( \Omega(x, \xi_q) a_{qj} + b_j \right)}
{\sum_{i=1}^K \sum_{q=1}^{KQ} \exp \left(\Omega(x, \xi_q) a_{qi} + b_i \right)}.
\end{equation}

\section{MNIST dataset}

In order to illustrate our approach we consider the well known MNIST data set, which is a large database of handwritten digits (0,1,...,9), containing 60,000 training images 
and 10,000 testing images \cite{key-6,key-7}. These are monochrome images with an intensity in the interval $[0,255]$ and the size of $28 \times 28$ pixels. 
The MNIST data set is probably the most frequently used benchmark in image recognition. 

In our first numerical experiment we only use the raw data, without any augmentation or distortion.
To our knowledge, the best reported results in the literature for the kernel SVM classification of the MNIST raw data have a classification error 
of $1.1\%$,\cite{key-8} and respectively $1.4\%$,\cite{key-9} and they have been obtained by combining ten kernel SVM classifiers. 

In a second experiment we "engineer" extra features by concatenating the images with the square root of the absolute value of their fast Fourier transform ($\text{FFT}$) \cite{key-4}. 
Since the FFT is symmetrical, only the first half of the values were used, such that each image becoming a vector of 1176 elements:
\begin{align}
x_i &\leftarrow x_i - \langle x_i \rangle \\
f_i &\leftarrow \vert \text{FFT}(x_i) \vert^{1/2}, \\
f_i &\leftarrow f_i - \langle f_i \rangle \\
x_i &\leftarrow x_i/\Vert x_i \Vert, \\
f_i &\leftarrow f_i/\Vert f_i \Vert, \\
x_i &\leftarrow \frac{1}{\sqrt{2}} [x_i, f_i]^T,
\end{align}
where $\langle . \rangle$ is the average.

In Fig. 1 we give the results for $100 \leq Q \leq 2500$. One can see that the classification error decreases from 
$\eta_{RAW} = 3.0 \%$ and $\eta_{RAW-FFT} = 2.9 \%$ for $Q=100$, to $\eta_{RAW} = 1.3 \%$ and $\eta_{RAW-FFT} = 1.1 \%$ for $Q=2500$. 
The results are averaged over $T=100$ runs, since the K-means initialization is a random process. 

\begin{figure}[h!]
\centering \includegraphics[width=7cm]{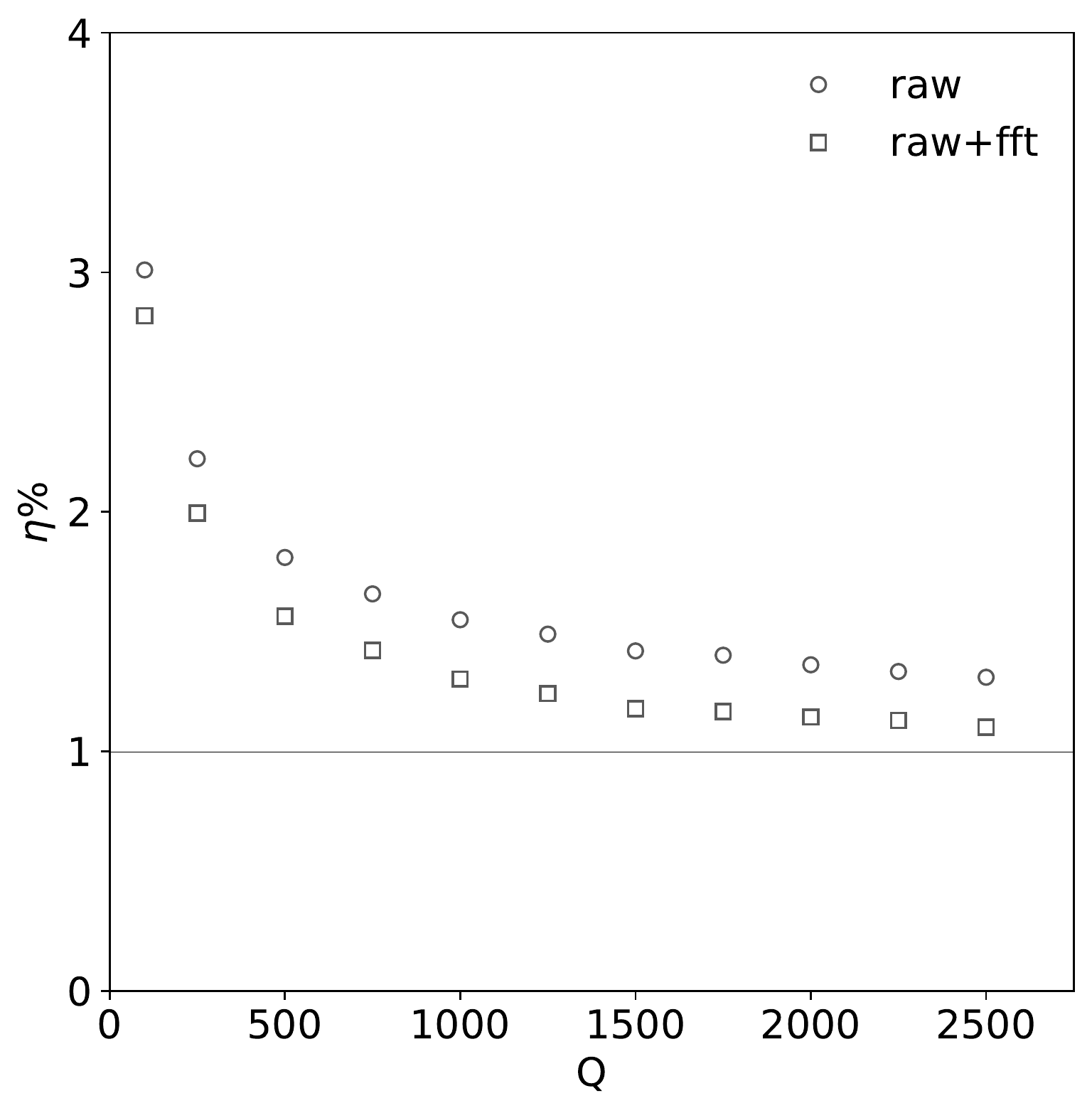}
\caption{The classification error $\eta$ as a function of the number of representative vectors $Q$ per class. 
}
\end{figure}

In a third experiment we used the raw data, from which we extracted a set of characteristic features for each class.  
In order to do so, we iterate over all images $x_n$ from a class $k$, and we extract all the overlapping patches (sub-images) of a fixed size $\ell \times \ell$, 
where $\ell < L=28$: $x_{nij} \subset x_n$, $i,j \in \{1,...,L-\ell\}$. 
The patches are vectorized by concatenating the columns, and then normalized as following:

\begin{align}
x_{nij} &\leftarrow x_{nij} - \langle x_{nij} \rangle, \\
x_{nij} &\leftarrow x_{nij}/\Vert x_{nij} \Vert. 
\end{align}
Thus, from each image we extract $(L-\ell+1) \times (L-\ell+1)$ patches. 
These patches are then used to define the "most common set of features" for each class. 
This is done by clustering the patches using K-means, such that for each class we extract $Q$ centroids features.
Finally with each centroid we associate the corresponding class label $C_k$ and we train the LS-SVM classifier. 

In order to classify a test image $\tilde{x}$ we extract all the patches: $\tilde{x}_{ij}$, $i,j \in \{1,...,L-\ell\}$. 
We classify each patch $\tilde{x}_{ij}$ using the LS-SVM classifier, and in the end we apply the simple majority rule 
to decide the class of the test sample. That is, we say that the the sample $\tilde{x}\in C_k$ if the majority of its patches 
are classified in the class $C_k$. Here we report the results obtained for $\ell=25$. Thus, from each image we extract 16 patches, and therefore the total number of 
training patches is 960,000. The results for $100 \leq Q \leq 5000$ are given in Fig. 2. One can see that in this case the error drops to $\varepsilon = 0.87\%$, 
which is a significant improvement over the previous results.

\begin{figure}[h!]
\centering \includegraphics[width=7cm]{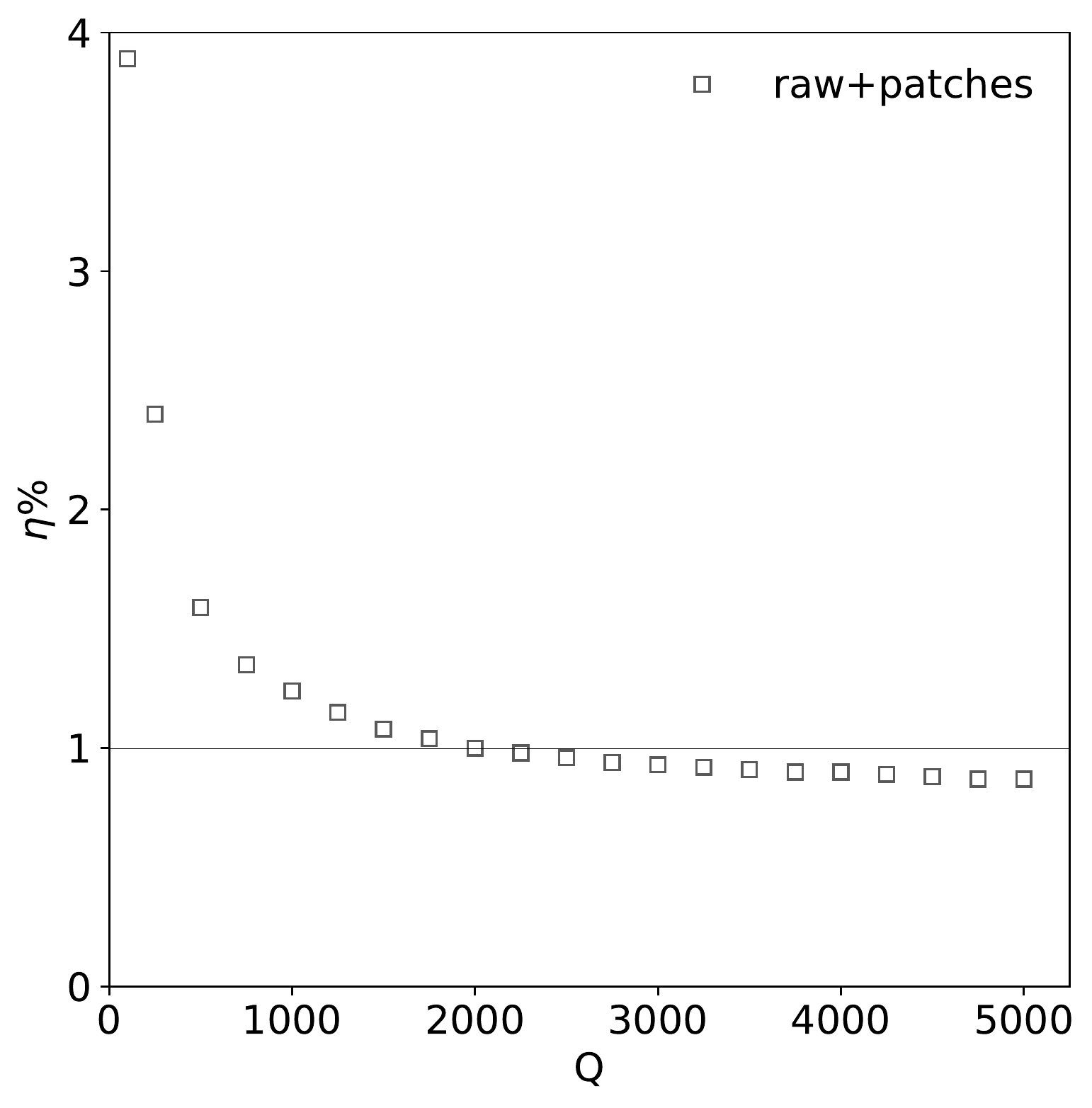}
\caption{The classification error $\eta$ as a function of the number of representative patch vectors $Q$ per class. 
}
\end{figure}

\section{Numerical implementation}

Our numerical implementation is in Python, and uses the "numpy" and and "scipy" modules. 
The K-means algorithm is reformulated such that it can exploit optimally the Intel's MKL BLAS implementation via "numpy" and "scipy". 
More exactly, the version of the K-means algorithm which we developed here is based on the cosine similarity measure, and uses only matrix-matrix multiplications (dense/sparse, accordingly) 
which are BLAS optimal. This is important, since we assume all the time that the samples and the centroids are normalized such that their 
Euclidean norm is one. 

Assuming that $\{x^{k}_i \vert x^{k}_i \in \mathbb{R}^M, i=1,\dots,N_k\}$ are the $N_k$ training samples, 
and $\{\xi^{k}_q \vert \xi^{k}_q \in \mathbb{R}^M, q=1,\dots,Q\}$ are the $Q<N_k$ centroids, we proceed as following: 

\begin{enumerate}

\item We build the matrix $X_k=[x^{k}_{im}]_{N_k \times M}$, where each row is a training sample from class $C_k$. 

\item We build the matrix $\Xi_k=[\xi^{k}_{qm}]_{Q \times M}$, where each row is a centroid initialized with a randomly chosen training sample. 

\item We take the dense matrix-matrix product:  
\begin{equation}
R_k = X_k \Xi_k^T = [r^{k}_{iq}]_{N_k \times Q}.
\end{equation}

\item The matrix $R_k$ is transformed into a sparse matrix $\hat{R}_k=[\hat{r}^k_{iq}]$, where:
\begin{equation}
\hat{r}^k_{iq} = 
\begin{cases}
		1 & \text{if }\; q=\text{arg}\max_q r^k_{iq} \\
		0 & \text{otherwise}
\end{cases} \quad i=1,...,N_k,
\end{equation}
this way each sample is assigned to the most similar centroid.

\item We take the sparse matrix-matrix product, to obtain a new set of centroids:
\begin{equation}
\hat{\Xi}_k = \hat{R}^T_k X_k = [\hat{\xi}^{k}_{qm}]_{Q \times M}.
\end{equation}

\item We normalize the new set of centroids:
\begin{equation}
\hat{\xi}^k_{q} \leftarrow \hat{\xi}^k_q/\Vert \hat{\xi}^k_q \Vert, \quad q=1,...,Q.
\end{equation}

\item We compute the alignment deviation between the new set and the old set of centroids:
\begin{equation}
\delta = 1 - \frac{1}{Q}\sum_{q=1}^Q \langle \hat{\xi}^k_{q}, \xi^k_{q} \rangle.
\end{equation}

\item We copy the new centroids matrix set into the old one: $\Xi_k \leftarrow \hat{\Xi}_k$. 

\item If $\delta > \tau$ then go to Step 3, otherwise return $\Xi_k$. Here $\tau$ is a small acceptable threshold ($\tau=10^{-6}$ in our implementation). 

\end{enumerate}

The LS-SVM also can be implemented in just a few highly efficient lines of Python code. 
We have experimented with several kernel types (Gaussian, polynomial), and the best results have been obtained with 
a polynomial kernel of degree four:
\begin{equation}
\Omega(x,x') = \langle x, x' \rangle^4.
\end{equation}
Therefore all the results reported here are for this particular kernel function. Also, the regularization parameter was always set to $\varepsilon= 10^{-6}$, 
and the classification error $\eta$ was simply measured as the percentage of the test images which have been incorrectly classified.

\section*{Conclusion}
In this paper we have combined the K-means clustering algorithm with the LS-SVM kernel classification method. 
We have shown how K-means clustering can be used to extract a set of representative vectors for each class, 
and how the LS-SVM kernel method uses this set of representative vectors as a new training set to perform the required classification task. 
The described approach has a couple of significant advantages over the previously described methods: 
(1) it is extremely robust, since it a single tuning parameter (the number of support vectors per class), 
and therefore data overfitting is easily avoided; (2) it is very simple to implement comparing to the previously proposed randomization methods. 
A full implementation of the method is also provided in \cite{key-10}, and it is illustrated with the popular MNIST dataset.

\end{document}